\documentclass{IJCAS}

\usepackage{url}
\usepackage{array,tabularx}
\usepackage{multicol} 
\hypersetup{hidelinks}

\journalvolumn{VV}
\journalnumber{X}
\journalyear{YYYY}
\setarticlestartpagenumber{1}

\allowdisplaybreaks

\begin{document}
\title{Safety-Critical Whole-Body Control for Humanoid Robots via Input-to-State Safe Control Barrier Functions}

\author{Kwanwoo Lee, Sanghyuk Park, Gyeongjae Park, Myeong-Ju Kim, and Jaeheung Park*}


\begin{abstract}
Safety-critical control is essential for humanoid robots operating in complex human-centered environments, where physical safety constraints such as joint limits, self-collision avoidance, obstacle avoidance, and workspace boundaries must be satisfied during real-robot operation.
However, existing approaches remain limited because kinematic safety guarantees can be degraded in the presence of unknown disturbances, such as model uncertainties, trajectory-tracking errors, and external perturbations.
This paper presents a hierarchical safety-critical whole-body control framework for humanoid robots based on input-to-state safe control barrier functions (ISSf-CBFs). 
The proposed architecture integrates a kinematic-level whole-body controller (KinWBC), an ISSf-CBF safety filter, and a dynamic-level whole-body controller (DynWBC). 
KinWBC generates nominal joint-motion references from prioritized tasks; the ISSf-CBF filter minimally modifies these references to satisfy kinematic safety constraints under bounded disturbances; and DynWBC tracks the filtered references while enforcing full-body dynamic feasibility and contact stability.
Safety constraints are imposed on a whole-body kinematic model, and the ISSf-CBF parameters are conservatively tuned so that the resulting kinematic safety guarantees can be transferred to full-order humanoid dynamics under unknown disturbances.
Simulation and real-robot experiments demonstrate that the proposed framework improves safety margins under model mismatch and reliably enforces multiple safety constraints in real time during locomotion, teleoperation, and single-leg balancing with hand control.

Project website: \url{https://kwlee365.github.io/SafeWBC-Website/}
\end{abstract}


\maketitle

\makeAuthorInformation{
Kwanwoo Lee, Sanghyuk Park, Gyeongjae Park, and Jaeheung Park are with the Department of Intelligence and Information, Seoul National University, Republic of Korea (e-mails: \{kwlee365, sang0823, rudwo1301, park73\}@snu.ac.kr). 
Myeong-Ju Kim is with the Robotics Lab, Hyundai Motor Group, Republic of Korea (e-mail: myeong-ju@snu.ac.kr). 
Jaeheung Park is also with the Advanced Institute of Convergence Technology, Republic of Korea, and with ASRI, AIIS, Seoul National University, Republic of Korea. 
He is the corresponding author of this paper (e-mail: park73@snu.ac.kr).

* Corresponding author.
}

\runningtitle{2025}{Kwanwoo Lee, Sanghyuk Park, Gyeongjae Park, Myeong-Ju Kim, and Jaeheung Park}{Manuscript Template for the International Journal of Control, Automation, and Systems: ICROS {\&} KIEE}{xxx}{xxxx}{x}

\section{INTRODUCTION}

\subsection{Motivation}
Humanoid robots have attracted increasing attention as versatile platforms capable of performing diverse tasks in complex, human-centered environments. 
To operate safely in such environments, humanoid robots must perform tasks while simultaneously accounting for hardware-related constraints, including joint limits, self-collision avoidance, and workspace boundaries, as well as safety constraints arising from robot-environment interactions, such as collision avoidance with nearby objects.
Unsafe control actions may lead to hazardous interactions with the robot itself or nearby objects; therefore, safety-critical control is a fundamental requirement for real-world deployment. 
Existing approaches commonly rely on safety filters to modify nominal motions so that the resulting motions satisfy safety constraints. 
However, in practice, robotic systems are often subject to unknown disturbances, such as model uncertainties, trajectory-tracking errors, and external perturbations; thus, safety guarantees established at the kinematic level may be degraded during real-robot operation.

This work aims to address this limitation by consistently transferring kinematically safe motion references to the full-order dynamics level, thereby achieving both safety assurance and task performance even in the presence of unknown disturbances.
\subsection{Related Work}
Two representative approaches for enforcing safety constraints in whole-body control are Artificial Potential Fields (APFs)~\cite{oussama1986apf} and Control Barrier Functions (CBFs)~\cite{ames2019cbf}.
APF-based methods enforce safety by generating repulsive control actions from the gradient of an artificial potential defined near constraint boundaries.
As the robot approaches a boundary, this potential can vary sharply, leading to large changes in its gradient.
Because this gradient is mapped directly into the control input, APF-based controllers may exhibit oscillatory behavior near constraint boundaries~\cite{koren1991potential, singletary2021comparative}.
Moreover, when multiple objectives and constraints must be handled simultaneously, APF-based frameworks often rely on explicit hierarchical projections, which can limit scalability as the number of constraints increases~\cite{oussama2022constraint}.
In contrast, CBFs enforce safety through optimization-based inequality constraints rather than by directly injecting repulsive gradients into the control input~\cite{ames2017cbf_qp}.
This yields a minimally invasive safety filter that modifies the nominal command only when necessary while preserving task performance as much as possible.
Accordingly, safety-critical whole-body control using CBFs has been demonstrated on a variety of robotic systems, including manipulators~\cite{murtaza2021torquesaturation, Kurtz2021singularity, Ko2024jerk, morton2025safe}, quadrupedal robots~\cite{grandia2021multi, kim2023coordination}, and humanoid robots~\cite{khazoom2022collision, safe2024paredes}.

For CBF-based methods, safety can generally be ensured either through velocity-based inverse kinematics or acceleration-based inverse dynamics.
Velocity-based inverse kinematics enforces safety in whole-body control by imposing CBF constraints at the joint-velocity level~\cite{singletary2022food, singletary2022kinematic, ding2024cbf}.
However, velocity-based inverse-kinematics controllers cannot explicitly account for full-body dynamics or enforce contact stability constraints. 
For these reasons, they may have limited applicability to legged robots, where contact stability must be explicitly considered to achieve stable locomotion.
On the other hand, acceleration-based whole-body inverse dynamics controllers are generally preferred for dynamic locomotion because they explicitly account for full-body dynamics and contact stability constraints.
Recent studies have incorporated exponential control barrier functions (eCBFs)~\cite{nguyen2022exponential} into such controllers to enforce kinematic safety constraints, such as joint limits and self-collision avoidance, during humanoid locomotion~\cite{khazoom2022collision, safe2024paredes}.
However, because the safety guarantees of such approaches inherently rely on high-fidelity dynamic models and accurate tracking of kinematic references by low-level controllers, safety constraints can be violated in the presence of model uncertainties or tracking errors~\cite{xu2015robustness, nguyen2022robust}.

In recent years, researchers have increasingly adopted input-to-state safe control barrier functions (ISSf-CBFs) to ensure safety for nonlinear dynamical systems in the presence of input disturbances~\cite{kolathaya2019issf}.
Specifically, ISSf-CBFs generalize safety by establishing forward invariance of a slightly enlarged safe set, ensuring that the state remains inside or close to the original safe set, with the deviation bounded by the disturbance magnitude.
Building on this concept, several studies have investigated the integration of ISSf-CBFs with reduced-order models (ROMs) to facilitate safety-critical controller design for complex, high-dimensional systems~\cite{cohen2024safety, molnar2022modelfree, molnar2023safety, cohen2025safety}.
In this line of work, safety constraints are formulated on simplified ROMs that generate safe reference commands, while the discrepancy between the ROM and the corresponding full-order model (FOM) is modeled as an input disturbance. 
ISSf-CBFs are then used to ensure that the safety guarantees established on the ROM are effectively transferred to the FOM.
Such frameworks have demonstrated promising results in applications including autonomous driving~\cite{alan2023issfcbf}, quadrupedal robot navigation~\cite{molnar2023safety}, and 3D hopping robots~\cite{cohen2025safety}.
However, to the best of our knowledge, these approaches have not yet been extended to humanoid robots.

\subsection{Contribution}



This paper presents a hierarchical safety-critical whole-body control framework for humanoid robots based on ISSf-CBFs. 
The proposed ISSf-CBF safety filter modifies nominal joint-velocity references into kinematically safe references that satisfy the prescribed safety constraints.
By transferring these kinematically safe references to the full-order dynamics level, the proposed framework aims to achieve both safety assurance and task performance even in the presence of unknown disturbances during real-robot operation.
In contrast to approaches that rely only on whole-body inverse dynamics controllers~\cite{khazoom2022collision, safe2024paredes}, the proposed framework does not place all safety enforcement at the dynamic-control level, where safety assurance can depend heavily on high model fidelity and may be degraded by unknown disturbances.
Building on the multi-layered whole-body control architecture in~\cite{kim2020dynamic}, we introduce an ISSf-CBF-based safety filter between the kinematic-level and dynamic-level whole-body controllers.
The main contributions of this study are summarized as follows:
\begin{itemize}
    \item A hierarchical safety-critical whole-body control framework that jointly addresses kinematic safety, dynamic feasibility, and contact stability.

    \item A safety-transfer strategy from velocity-level kinematics to full-order dynamics that combines an ISSf-CBF-based safety filter with conservative parameter tuning, thereby preserving safety guarantees under bounded disturbances.

    \item Validation through simulations and real-robot experiments involving locomotion, teleoperation, and single-leg balancing with hand control, demonstrating reliable safety enforcement under model mismatch across diverse scenarios.
\end{itemize}

\subsection{Paper organization}
The remainder of this paper is organized as follows.
Section~\ref{sec:cbf} presents the notion of ISSf-CBFs.
Section~\ref{sec:wbc} describes the overall whole-body control architecture.
Section~\ref{sec:safety_filter} details the ISSf-CBF-based safety filter.
Section~\ref{sec:result} reports the simulation and experimental results.
Finally, Section~\ref{sec:conclusion} concludes the paper.

%
\section{Input-to-State Safe Control Barrier Functions} \label{sec:cbf}
In this section, we present the notion of safety and the formal guarantees provided by CBFs~\cite{ames2019cbf} as preliminaries for the ISSf-CBF-based safety filter proposed in Section~\ref{sec:safety_filter}.

Consider the control-affine system
\begin{equation}
    \dot{\bm{x}} = \bm{f}(\bm{x}) + \bm{g}(\bm{x})\bm{u},
    \label{eq:system}
\end{equation}
where $\bm{x} \in X \subseteq \mathbb{R}^n$ and $\bm{u} \in U \subseteq \mathbb{R}^m$ denote the system state and control input, respectively.
The functions $\bm{f} : X \rightarrow \mathbb{R}^n$ and $\bm{g} : X \rightarrow \mathbb{R}^{n \times m}$ are assumed to be locally Lipschitz continuous.
Given an initial condition $\bm{x}(0)=\bm{x}_0 \in X$ and a locally Lipschitz continuous controller $\bm{u} = \bm{k}(\bm{x})$, we assume that the closed-loop system admits a unique solution $\bm{x}(t)$ for all $t \ge 0$.
\subsection{Safety and Control Barrier Functions}

\textbf{Safe set:}
The set $S$ is forward invariant under~\eqref{eq:system} if $\bm{x}_0 \in S \Rightarrow \bm{x}(t) \in S$ for all $t \ge 0$.
We define the safe set as the superlevel set of a continuously differentiable function $h : X \rightarrow \mathbb{R}$:
\begin{equation}
    S \triangleq \{ \bm{x} \in X \mid h(\bm{x}) \ge 0 \}.
    \label{eq:safe}
\end{equation}
The system~\eqref{eq:system} is said to be \emph{safe} if its state remains in a prescribed \emph{safe set} $S \subset X$ for all $t \ge 0$.

\textbf{Control Barrier Function (CBF):}
A continuously differentiable function $h: X \rightarrow \mathbb{R}$ is a Control Barrier Function (CBF) for the set $S$
if there exists a class-$\mathcal{K}$ function $\alpha(\cdot)$ such that
\begin{equation}
    \sup_{\bm{u} \in U} \ \dot{h}(\bm{x},\bm{u}) \ge -\alpha(h(\bm{x})), \quad \forall \bm{x} \in \mathbb{R}^n,
\end{equation}
where
\begin{equation}
    \dot{h}(\bm{x},\bm{u}) = L_f h(\bm{x}) + L_g h(\bm{x})\bm{u}.
\end{equation}
Here, $L_f h(\bm{x}) = \frac{\partial h}{\partial \bm{x}} \bm{f}(\bm{x})$ and $L_g h(\bm{x}) = \frac{\partial h}{\partial \bm{x}} \bm{g}(\bm{x})$ denote the Lie derivatives of $h$ along the vector fields $\bm{f}$ and $\bm{g}$, respectively, so that $\dot{h}$ is the derivative of $h$ along the system dynamics.

If $h$ is a CBF, then any locally Lipschitz continuous controller $\bm{u} = \bm{k}(\bm{x})$ satisfying
\begin{equation}
    \dot{h}(\bm{x},\bm{k}(\bm{x})) \ge -\alpha(h(\bm{x})), \quad \forall \bm{x} \in \mathbb{R}^n,
    \label{eq:CBF_safety}
\end{equation}
renders the system~\eqref{eq:system} safe with respect to $S$.

\textbf{CBF-based Quadratic Program (CBF-QP):}
The condition~\eqref{eq:CBF_safety} enables safety-critical controller synthesis.
Given a nominal controller $\bm{k}_d(\bm{x})$, a minimally invasive safe controller $\bm{k}_s(\bm{x})$ can be obtained by solving
\begin{equation}
\begin{aligned}
    \bm{k}_s(\bm{x}) = \arg\min_{\bm{u} \in U} \ & \|\bm{u} - \bm{k}_d(\bm{x})\|^2 \\
    \text{s.t.} \ & \dot{h}(\bm{x},\bm{u}) \ge -\alpha(h(\bm{x})). 
\end{aligned}
\end{equation}
The resulting control law $\bm{k}_s(\bm{x})$ is Lipschitz continuous under standard regularity conditions~\cite{xu2015robustness, ames2017cbf_qp}.

\subsection{Input-to-State Safe Control Barrier Functions}
\label{sec:disturbance}

The above safety guarantees assume that the control input $\bm{u}$ is applied exactly.
In practice, robotic systems are subject to bounded input disturbances $\bm{d} \in \mathbb{R}^m$, resulting~in
\begin{equation}
    \dot{\bm{x}} = \bm{f}(\bm{x}) + \bm{g}(\bm{x})(\bm{u} + \bm{d}).
    \label{eq:disturbed_dynamics}
\end{equation}

\textbf{Input-to-State Safety (ISSf):}
Under disturbances, strict forward invariance of $S$ may not be preserved.
Instead, safety is characterized in the sense of input-to-state safety (ISSf)~\cite{kolathaya2019issf},
which ensures invariance of a disturbance-dependent enlargement of the safe set.
More precisely, this disturbance-dependent enlarged safe set is defined as
\begin{equation}
    S_d \triangleq \{ \bm{x} \in X \mid h(\bm{x}) + \gamma(\|\bm{d}\|_\infty) \ge 0 \},
\end{equation}
where $\gamma(\cdot)$ is a class-$\mathcal{K}$ function.
ISSf is achieved if $\bm{x}(0) \in S_d \Rightarrow \bm{x}(t) \in S_d$ for all $t \ge 0$.

\textbf{Input-to-State Safe Control Barrier Function (ISSf-CBF):}
A continuously differentiable function $h : X \rightarrow \mathbb{R}$ is called an Input-to-State Safe Control Barrier Function (ISSf-CBF) if there exist class-$\mathcal{K}$ functions $\alpha(\cdot)$ and $\iota(\cdot)$ such that
\begin{equation}
    \sup_{\bm{u} \in U} \ \dot{h}(\bm{x},\bm{u},\bm{d})
    \ge -\alpha(h(\bm{x})) - \iota(\|\bm{d}\|_\infty), 
    \quad
    \forall \bm{x} \in \mathbb{R}^n,
\end{equation}
where
\begin{equation}
    \dot{h}(\bm{x},\bm{u},\bm{d}) = 
        L_f h(\bm{x}) + L_g h(\bm{x})(\bm{u} + \bm{d}),
\end{equation}
for all bounded disturbances $\bm{d}$.

A sufficient condition for $h$ to satisfy the ISSf-CBF property can be obtained via a constructive argument.
Given a baseline controller $\bm{k}(\bm{x})$, consider the modified control input
\begin{equation}
    \bm{u}(\bm{x}) = \bm{k}(\bm{x}) + L_g h(\bm{x})^\top.
\end{equation}
If there exists a class-$\mathcal{K}$ function $\alpha(\cdot)$ such that
\begin{equation}
\begin{aligned}
    \sup_{\bm{u} \in U}  \ \Big(
        L_f h(\bm{x}) + L_g h(\bm{x})\bm{u} - L_g h(\bm{x})L_g h(\bm{x})^\top
    \Big)
    &\ge -\alpha(h(\bm{x})), \\
    &\forall \bm{x} \in \mathbb{R}^n,
\end{aligned}
\label{eq:issf_cbf_sufficient}
\end{equation}
then $h$ is an ISSf-CBF, and the safe set is rendered input-to-state safe under bounded disturbances.
The above sufficient condition follows from a constructive ISSf-CBF design, in which the disturbance-dependent term is explicitly upper-bounded by completing the square in the barrier derivative.
The detailed proof can be found in~\cite{kolathaya2019issf}.

\section{Overall Whole-Body Control Architecture} \label{sec:wbc}
\begin{figure*}[!t]
    \centering
    \includegraphics[width=\textwidth]{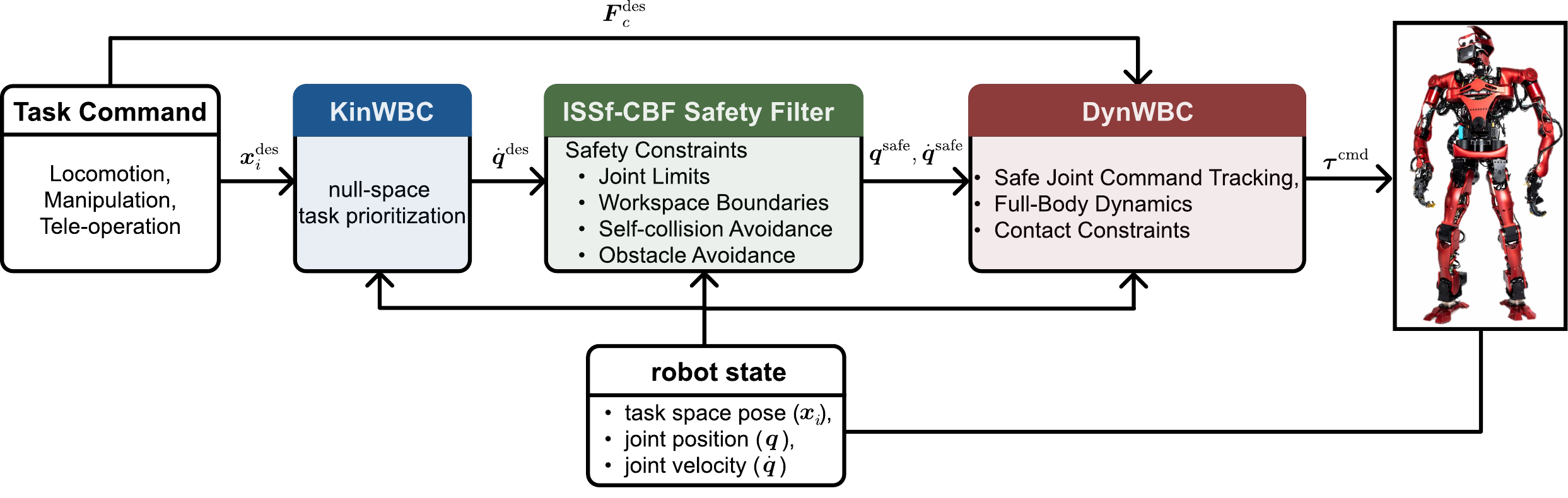}
    \caption{Overall hierarchical whole-body control architecture of the proposed framework.}
    \label{fig:block_diagram}
\end{figure*}

This section presents the hierarchical whole-body control architecture, building on the framework in~\cite{kim2020dynamic}. 
Fig.~\ref{fig:block_diagram} illustrates the overall control architecture.

\subsection{Kinematic-level whole-body controller (KinWBC)}
Given a set of prioritized operational-space tasks (e.g., locomotion, manipulation, and teleoperation), KinWBC computes joint-position references using hierarchical differential inverse kinematics. Its primary role is to coordinate multiple tasks through null-space projections, ensuring that lower-priority tasks are achieved without degrading the tracking performance of higher-priority tasks.

Let the $i$-th task be defined by its task Jacobian $\bm{J}_i$ and desired task-space command $\bm{x}_i^{\text{des}}$.
KinWBC iteratively updates the joint-velocity command $\dot{\bm{q}}^{\text{des}}$ using a prioritized pseudo-inverse in the projected null space:
\begin{equation}
\dot{\bm{q}}_i^{\text{des}} = \dot{\bm{q}}_{i-1}^{\text{des}} + \bm{J}_{i|\text{pre}}^{\dagger}\left( \bm{x}_i^{\text{des}} - \bm{x}_i - \bm{J}_i \dot{\bm{q}}_{i-1}^{\text{des}} \right),
\label{eq:kinwbc_qdot}
\end{equation}
where $\bm{J}_{i|\text{pre}} \triangleq \bm{J}_i \bm{N}_{i-1}$ denotes the Jacobian projected onto the null space of higher-priority tasks, $\bm{N}_{i-1}$ is the corresponding null-space projector, and $(\cdot)^{\dagger}$ denotes a singular value decomposition (SVD)-based pseudo-inverse operator in which small singular values are set to 0. 
The recursion is initialized with $\dot{\bm{q}}_0^{\text{des}}=\bm{0}$ and $\bm{N}_0=\bm{I}$.
Then, the desired joint position~$\bm{q}^{\text{des}}$ is obtained by integration as
\begin{equation}
\bm{q}^{\text{des}} = \bm{q} + \dot{\bm{q}}^{\text{des}}\,dt,
\label{eq:kinwbc_qdes}
\end{equation}
where $\bm{q}$ represents the full generalized configuration, including the floating base and actuated joints, and $dt$ is the integration step size.

In~\cite{kim2020dynamic}, the desired joint position $\bm{q}^{\text{des}}$ and velocity $\dot{\bm{q}}^{\text{des}}$ are mapped to a desired joint-acceleration command for DynWBC.
In contrast, the proposed framework applies an ISSf-CBF-based safety filter to the joint-velocity command, yielding a filtered velocity command $\dot{\bm{q}}^{\text{safe}}$.
The safety-filtered kinematic reference is constructed as $\bm{q}^{\text{safe}} \triangleq \bm{q} + \dot{\bm{q}}^{\text{safe}}dt$ and is subsequently converted into the desired joint-acceleration command via the feedback law
\begin{equation}
\ddot{\bm{q}}^{\text{safe}}
=
\bm{K}_{p}^{\text{dyn}}(\bm{q}^{\text{safe}}-\bm{q})
+
\bm{K}_{d}^{\text{dyn}}(\dot{\bm{q}}^{\text{safe}}-\dot{\bm{q}}),
\label{eq:qddot_safe_feedback}
\end{equation}
where $\bm{K}_{p}^{\text{dyn}}$ and $\bm{K}_{d}^{\text{dyn}}$ are feedback gain matrices.

\subsection{Dynamic-level whole-body controller (DynWBC)}
DynWBC solves a QP that tracks the safe acceleration reference $\ddot{\bm{q}}^{\text{safe}}$ while jointly optimizing the contact reaction wrenches and joint torques subject to full-body dynamics and contact wrench cone constraints.
Let $\bm{F}_c$, $\ddot{\bm{q}}$, and $\bm{\tau}$ denote the stacked contact reaction wrenches at the left and right feet, the generalized acceleration, and the joint torque, respectively.
Given desired quantities $\ddot{\bm{q}}^{\text{safe}}$ and $\bm{F}_c^{\text{des}}$, DynWBC solves
\begin{align}
\min_{\bm{F}_c,\,\ddot{\bm{q}},\,\bm{\tau}}\quad
\label{eq:dynwbc_qp_obj}
& w_{\ddot{q}} \left\|\ddot{\bm{q}} - \ddot{\bm{q}}^{\text{safe}}\right\|^2
+ w_c \left\|\bm{F}_c - \bm{F}_c^{\text{des}}\right\|^2 
\nonumber \\
&\quad + w_\tau \left\|\bm{\tau}-\bm{\tau}_{\text{prev}}\right\|^2
+ w_M \, \ddot{\bm{q}}^\top \bm{M}(\bm{q}) \ddot{\bm{q}},
\\
\text{s.t.}\quad
& \bm{U}\bm{F}_c \le \bm{0},
\label{eq:dynwbc_qp_fric}
\\
& \bm{M}(\bm{q})\,\ddot{\bm{q}} + \bm{h}(\bm{q},\dot{\bm{q}})
= \bm{S}^\top \bm{\tau} + \bm{J}_c(\bm{q})^\top \bm{F}_c.
\label{eq:dynwbc_qp_dyn}
\end{align}

The objective in~\eqref{eq:dynwbc_qp_obj} consists of four weighted terms.
The first term penalizes the tracking error of the generalized acceleration $\ddot{\bm{q}}$ with respect to $\ddot{\bm{q}}^{\text{safe}}$, weighted by $w_{\ddot{q}}$, thereby prioritizing acceleration-level tracking of the safety-filtered reference.
The second term penalizes deviations of the contact reaction wrenches $\bm{F}_c$ from the reference $\bm{F}_c^{\text{des}}$, weighted by $w_c$, to track desired contact wrenches.
Here, $\bm{F}_c^{\text{des}}$ can be computed to maintain balance during locomotion and manipulation, for example, using a model predictive controller (MPC)~\cite{di2018dynamic} or a simple contact wrench distributor~\cite{kajita2010biped}.
The third term penalizes changes in the commanded joint torque relative to the previous command $\bm{\tau}_{\text{prev}}$, weighted by $w_\tau$, which promotes smooth variation of torque commands over time.
Finally, the term $\ddot{\bm{q}}^\top \bm{M}(\bm{q}) \ddot{\bm{q}}$, weighted by $w_M$, regularizes acceleration energy as defined in~\cite{bruyninckx2000gauss} and helps suppress aggressive whole-body accelerations.

The constraints consist of two components.
First,~\eqref{eq:dynwbc_qp_fric} enforces contact stability via linear inequalities~\cite{caron2015contactwrenchcone}, where $\bm{U}$ encodes a linearized 6D contact wrench cone including unilateral normal force (no lift-off), Coulomb friction (no slip), and bounded contact moments (e.g., center-of-pressure and torsional limits).
Second,~\eqref{eq:dynwbc_qp_dyn} imposes the humanoid full-body equations of motion, where $\bm{M}(\bm{q})$ denotes the generalized inertia matrix, $\bm{h}(\bm{q},\dot{\bm{q}})$ collects Coriolis, centrifugal, and gravitational terms, $\bm{S}$ selects actuated joints, and $\bm{J}_c(\bm{q})$ is the contact Jacobian.

Finally, the motor torque command applied to the actuators is
\begin{equation}
\bm{\tau}^{\mathrm{cmd}} = \bm{\tau}^{\mathrm{opt}} + \bm{K}_p(\bm{q}^{\text{safe}}-\bm{q}) + \bm{K}_d(\dot{\bm{q}}^{\text{safe}}-\dot{\bm{q}}),
\label{eq:motor_torque_command}
\end{equation}
where the first term $\bm{\tau}^{\mathrm{opt}}$ is the optimal torque input computed from \eqref{eq:dynwbc_qp_obj}, and the remaining terms introduce motor PD feedback to help overcome joint friction and other unmodeled disturbances.

\section{ISSf-CBF-BASED SAFETY FILTER} \label{sec:safety_filter}
This section presents the proposed ISSf-CBF-based safety filter, which modifies the joint-velocity references generated by KinWBC so that they satisfy prescribed kinematic safety constraints before being converted into the desired joint-acceleration command for DynWBC. 
Through this intermediate layer, kinematically safe joint-motion references are conservatively transferred to the full-order dynamics level, enabling robust safety assurance during real-robot operation.

\subsection{Reduced-order models (ROMs)}
To enforce safety at the kinematic level while accounting for discrepancies between kinematic references and full-order dynamics, we adopt a safety-critical controller synthesis framework based on reduced-order models (ROMs)~\cite{cohen2025safety}.
In this work, safety constraints are imposed on an idealized ROM, and the resulting safety guarantees are transferred to the corresponding full-order model (FOM) via ISSf-CBFs.

To formalize this idea, the full-order state and input are defined as
$\bm{x}\triangleq
\begin{bmatrix}
\bm{q}^\top &
\dot{\bm{q}}^\top
\end{bmatrix}^\top$
and $\bm{u}\triangleq \bm{\tau}$, respectively.
The corresponding FOM is written as
\begin{equation}
\dot{\bm{x}} = \bm{F}(\bm{x},\bm{u})
\triangleq
\begin{bmatrix}
\dot{\bm{q}}\\[2pt]
\bm{M}(\bm{q})^{-1}\!\Big(
\bm{S}^\top \bm{\tau}
+ \bm{J}_c(\bm{q})^\top \bm{F}_c
- \bm{h}(\bm{q},\dot{\bm{q}})
\Big)
\end{bmatrix}.
\label{eq:fom_general}
\end{equation}

To obtain a reduced-order representation of~\eqref{eq:fom_general}, a differentiable state projection map
$\pi(\cdot)$ is defined to map the full-order state to a reduced-order state as
\begin{equation}
\bm{y} \triangleq \pi(\bm{x}).
\label{eq:proj_maps}
\end{equation}
Here, $\bm{y}$ denotes the reduced-order state on which safety constraints are imposed.
The state projection $\pi(\cdot)$ induces the projected dynamics
\begin{equation}
\dot{\bm{y}}
=
\frac{\partial \pi}{\partial \bm{x}}(\bm{x})\,\bm{F}(\bm{x},\bm{u}).
\label{eq:projected_dyn}
\end{equation}
While~\eqref{eq:projected_dyn} is an exact reduced-space representation of the FOM, it still depends on the full-order state and full-order input, which complicates reduced-order controller synthesis.
Therefore, an idealized ROM characterized by locally Lipschitz functions $\bm{f}$ and $\bm{g}$ is introduced:
\begin{equation}
\dot{\bm{y}} = \bm{f}(\bm{y}) + \bm{g}(\bm{y})\,\bm{v}.
\label{eq:rom_ideal}
\end{equation}
Here, $\bm{f}(\bm{y})$ and $\bm{g}(\bm{y})$ characterize the ROM dynamics, and $\bm{v}$ denotes the reduced-order input. 
In our case, the ROM control input is the safety-filtered joint-velocity command, i.e., $\bm{v}=\dot{\bm{q}}^{\text{safe}}$, as defined in~\eqref{eq:issf_qp_ki}.

To relate the ROM to the FOM, it is assumed that the reduced-order input $\bm{v}$ is mapped to the full-order input through a tracking law
\begin{equation}
\bm{u} = \bm{K}(\bm{x},\bm{v}).
\label{eq:tracking_law}
\end{equation}
In our case, the FOM control input is the joint torque command, i.e., $\bm{u}=\bm{\tau}^{\mathrm{cmd}}$, where $\bm{\tau}^{\mathrm{cmd}}$ is given by~\eqref{eq:motor_torque_command}.
Under this tracking law, the closed-loop FOM becomes
\begin{equation}
\dot{\bm{x}} = \bm{F}\!\bigl(\bm{x},\bm{K}(\bm{x},\bm{v})\bigr),
\label{eq:fom_closed_loop}
\end{equation}
and the corresponding reduced-order dynamics are
\begin{equation}
\dot{\bm{y}}
=
\frac{\partial \pi}{\partial \bm{x}}(\bm{x})\,
\bm{F}\!\bigl(\bm{x},\bm{K}(\bm{x},\bm{v})\bigr)
=
\bm{f}(\bm{y}) + \bm{g}(\bm{y})\,\bm{v} + \bm{d},
\label{eq:rom_true}
\end{equation}
where the discrepancy term is defined as
\begin{equation}
\bm{d}
\triangleq
\frac{\partial \pi}{\partial \bm{x}}(\bm{x})\,
\bm{F}\!\bigl(\bm{x},\bm{K}(\bm{x},\bm{v})\bigr)
-\bm{f}(\bm{y})
-\bm{g}(\bm{y})\,\bm{v}.
\label{eq:rom_dyn}
\end{equation}
Therefore, $\bm{d}$ represents the discrepancy between the projected closed-loop FOM dynamics and the idealized ROM.
In the ROM, $\bm{d}$ is treated as a disturbance term and can also be interpreted as the residual tracking mismatch between the commanded reduced-order input and the realized reduced-order behavior of the closed-loop FOM.
If $\bm{d}=\bm{0}$, the idealized ROM exactly captures the projected dynamics of the closed-loop FOM.
However, a nonzero discrepancy $\bm{d}$ may lead to safety violations.
Accordingly, $\bm{d}$ can be modeled as an unknown but bounded disturbance that may arise from factors such as modeling errors, external perturbations, and imperfect tracking of the reduced-order command by the full-order system.

\subsection{ISSf-CBF safety conditions on the disturbed ROM}
Building on the ROM dynamics in~\eqref{eq:rom_true}, we impose safety constraints on the reduced-order state $\bm{y} \in \mathbb{R}^{n_y}$.
We define the safe set for the ROM as the superlevel set of a continuously differentiable function $h:\mathbb{R}^{n_y}\rightarrow\mathbb{R}$:
\begin{equation}
\mathcal{S} \triangleq \left\{\bm{y}\in\mathbb{R}^{n_y}\mid h(\bm{y})\ge 0\right\}.
\end{equation}
Applying the ISSf-CBF sufficient condition in~\eqref{eq:issf_cbf_sufficient} to the ROM dynamics~\eqref{eq:rom_true} yields the following ROM-level ISSf-CBF condition:
\begin{equation}
L_f h(\bm{y}) + L_g h(\bm{y})\,\bm{v}
\ge
-\alpha\!\left(h(\bm{y})\right)
+ \frac{1}{\epsilon}\big\|
\nabla h(\bm{y})\big\|^2,
\label{eq:issf_cbf_rom}
\end{equation}
where $\epsilon>0$ is a user-defined parameter that indirectly helps restrict $\mathcal{S}_d$ to a smaller region.
The last term in~\eqref{eq:issf_cbf_rom} is included to compensate for the bounded disturbance $\bm{d}$.
Equation~\eqref{eq:issf_cbf_rom} ensures forward invariance of an enlarged safe set $\mathcal{S}_d$.
Therefore, enforcing~\eqref{eq:issf_cbf_rom} yields a safety filter that remains robust to bounded ROM--FOM discrepancies.
It is important to note that, following the practical tuning guideline in~\cite[Remark~2]{cohen2025safety}, safety transfer from the ROM to the FOM can be achieved by initializing $\alpha$ and $\epsilon$ with small values and then gradually increasing them until adequate performance is obtained.

\subsection{ISSf-CBF formulation for safety-critical kinematic control}
For the proposed kinematic safety filter, we adopt a whole-body velocity kinematic model as the ROM, with the reduced-order state and input defined as
\begin{equation}
\bm{y} \triangleq \bm{q},
\qquad
\bm{v} \triangleq \dot{\bm{q}}^{\text{safe}}.
\label{eq:ki_state_input}
\end{equation}
Then, the disturbed ROM is given by the velocity kinematic model
\begin{equation}
\dot{\bm{y}} = \bm{v} + \bm{d},
\label{eq:ki_dyn}
\end{equation}
which corresponds to $\bm{f}(\bm{y})=\bm{0}$ and $\bm{g}(\bm{y})=\bm{I}$ in~\eqref{eq:rom_ideal}.
To construct the safety filter, each safety constraint is enforced through an ISSf-CBF condition of the form
\begin{equation}
\nabla h(\bm{q})^\top \dot{\bm{q}}
\ge
-\alpha\!\left(h(\bm{q})\right)
+ \frac{1}{\epsilon}\big\|
\nabla h(\bm{q})\big\|^2.
\label{eq:issf_cbf_constraint_template}
\end{equation}
For simplicity, the class-$\mathcal{K}$ function $\alpha(\cdot)$ is chosen to be linear, i.e., $\alpha(h)=\alpha h$ with $\alpha\in\mathbb{R}_{+}$. 
Substituting each safety constraint into~\eqref{eq:issf_cbf_constraint_template} yields explicit inequalities in~$\dot{\bm{q}}$.

Accordingly, the proposed ISSf-CBF-based safety filter is formulated as the following minimally invasive QP:
\begin{align}
\dot{\bm{q}}^{\text{safe}}
&=
\arg\min_{\dot{\bm{q}}}\; \left\|\dot{\bm{q}}-\dot{\bm{q}}^{\text{des}}\right\|^2
\label{eq:issf_qp_ki}
\\
\text{s.t.}\; 
&\text{Joint-limit constraints,}  \label{eq:issf_qp_ki_joint} \\
&\text{Self-collision-avoidance constraints,}   \label{eq:issf_qp_ki_self} \\
&\text{Object-collision-avoidance constraints,}  \label{eq:issf_qp_ki_object} \\
&\text{Workspace-boundary constraints,}   \label{eq:issf_qp_ki_workspace} 
\end{align}
where $\dot{\bm{q}}^{\text{des}}$ denotes the nominal joint-velocity command generated by KinWBC in~\eqref{eq:kinwbc_qdot}.
The kinematic safety constraints consist of four components.

\textbf{Joint-limit constraints~\eqref{eq:issf_qp_ki_joint}:}
For each joint $i$, define the lower and upper boundary functions as
\begin{equation}
h_{i}^{\mathrm{JL},\min}(\bm{q}) = q_i - q_{i}^{\min},
\qquad
h_{i}^{\mathrm{JL},\max}(\bm{q}) = q_{i}^{\max} - q_i.
\label{eq:joint_limit_barriers}
\end{equation}
Using~\eqref{eq:issf_cbf_constraint_template}, the corresponding velocity-level inequalities are
\begin{align}
\dot{q}_i
&\ge
-\alpha_i^{\mathrm{JL},\min}\!h_{i}^{\mathrm{JL},\min}(\bm{q})
+ \frac{1}{\epsilon_i^{\mathrm{JL},\min}},
\label{eq:joint_limit_qdot_min}
\\
-\dot{q}_i
&\ge
-\alpha_i^{\mathrm{JL},\max}\!h_{i}^{\mathrm{JL},\max}(\bm{q})
+ \frac{1}{\epsilon_i^{\mathrm{JL},\max}}.
\label{eq:joint_limit_qdot_max}
\end{align}

\textbf{Self-collision-avoidance constraints~\eqref{eq:issf_qp_ki_self}:}
For efficient geometric computation, each robot link is approximated by collision spheres and capsules.
For each pair of collision bodies $(A,B)$, define the signed distance as
\begin{equation}
h_{AB}^{\mathrm{SC}}(\bm{q})
=
\left\|\bm{p}_{A}^{r}(\bm{q})-\bm{p}_{B}^{r}(\bm{q})\right\|_2 - (\rho_A+\rho_B),
\label{eq:self_collision_barrier}
\end{equation}
where $\bm{p}_{A}^{r}$ and $\bm{p}_{B}^{r}$ denote the center of a sphere or the closest point on a capsule centerline segment for collision bodies $A$ and $B$, respectively, and $\rho_A$ and $\rho_B$ are their corresponding radii.
Let
\begin{equation}
\hat{\bm{n}}_{AB}(\bm{q})
\triangleq
\frac{\bm{p}_{A}^{r}(\bm{q})-\bm{p}_{B}^{r}(\bm{q})}{\left\|\bm{p}_{A}^{r}(\bm{q})-\bm{p}_{B}^{r}(\bm{q})\right\|_2},
\qquad
s_{AB}\in\{+1,-1\},
\label{eq:self_collision_normal}
\end{equation}
and
\begin{equation}
\bm{J}_{AB}(\bm{q})
\triangleq
\frac{\partial h_{AB}^{\mathrm{SC}}}{\partial \bm{q}}
=
s_{AB}\,\hat{\bm{n}}_{AB}(\bm{q})^\top\!\left(\bm{J}_{A}(\bm{q})-\bm{J}_{B}(\bm{q})\right),
\label{eq:self_collision_signed_jacobian}
\end{equation}
where $s_{AB}$ takes the sign of $h_{AB}^{\mathrm{SC}}(\bm{q})$, and
\begin{equation}
\bm{J}_{A}(\bm{q}) \triangleq \frac{\partial \mathrm{FK}_{\bm{p}_{A}}(\bm{q})}{\partial \bm{q}},
\qquad
\bm{p}_{A}=\bm{p}_{A}^{r}+\rho_A\hat{\bm{n}}_{AB}(\bm{q}),
\label{eq:self_collision_point_jacobian}
\end{equation}
with an analogous definition for body $B$.
Then,~\eqref{eq:issf_cbf_constraint_template} yields
\begin{equation}
\bm{J}_{AB}(\bm{q})\dot{\bm{q}}
\ge
-\alpha_{AB}^{\mathrm{SC}}\!h_{AB}^{\mathrm{SC}}(\bm{q})
+ \frac{1}{\epsilon_{AB}^{\mathrm{SC}}}
\left\|\bm{J}_{AB}(\bm{q})\right\|^2.
\label{eq:self_collision_qdot}
\end{equation}

\textbf{Object-collision-avoidance constraints~\eqref{eq:issf_qp_ki_object}:}
Similarly, the robot and obstacle geometries are approximated by spheres and capsules.
For each robot collision body $A$ and obstacle collision body $O$, define
\begin{equation}
h_{AO}^{\mathrm{OC}}(\bm{q})
=
\left\|\bm{p}_{A}^{r}(\bm{q})-\bm{p}_{O}^{r}\right\|_2 - (\rho_A+\rho_O),
\label{eq:object_collision_barrier}
\end{equation}
where $\bm{p}_{O}^{r}$ denotes either the center of an obstacle sphere or the closest point on the centerline segment of an obstacle capsule.
Let
\begin{equation}
\hat{\bm{n}}_{AO}(\bm{q})
\triangleq
\frac{\bm{p}_{A}^{r}(\bm{q})-\bm{p}_{O}^{r}}{\left\|\bm{p}_{A}^{r}(\bm{q})-\bm{p}_{O}^{r}\right\|_2},
\qquad
s_{AO}\in\{+1,-1\},
\label{eq:object_collision_normal}
\end{equation}
and
\begin{equation}
\bm{J}_{AO}(\bm{q})
\triangleq
s_{AO}\,\hat{\bm{n}}_{AO}(\bm{q})^\top\bm{J}_{A}(\bm{q}),
\qquad
v_{AO}
\triangleq
s_{AO}\,\hat{\bm{n}}_{AO}(\bm{q})^\top\bm{v}_O,
\label{eq:object_collision_signed_jacobian}
\end{equation}
where $s_{AO}$ takes the sign of $h_{AO}^{\mathrm{OC}}(\bm{q})$, and $\bm{J}_{A}(\bm{q})$ is defined as in~\eqref{eq:self_collision_point_jacobian} with $\hat{\bm{n}}_{AB}$ replaced by $\hat{\bm{n}}_{AO}$.
Here, $\bm{v}_O$ denotes the obstacle point velocity, estimated using a linear Kalman filter~\cite{przybyla2017detection}.
Then,~\eqref{eq:issf_cbf_constraint_template} yields
\begin{equation}
\bm{J}_{AO}(\bm{q})\dot{\bm{q}}
\ge
v_{AO}
-\alpha_{AO}^{\mathrm{OC}}\!h_{AO}^{\mathrm{OC}}(\bm{q})
+ \frac{1}{\epsilon_{AO}^{\mathrm{OC}}}
\left\|\bm{J}_{AO}(\bm{q})\right\|^2.
\label{eq:object_collision_qdot}
\end{equation}

\textbf{Workspace-boundary constraints~\eqref{eq:issf_qp_ki_workspace}:}
For a designated point pair $(A,B)$ (e.g., shoulder and hand points), define
\begin{equation}
h^{\mathrm{WS}}_{\text{AB}}(\bm{q})
=
d_{\max} - \|\bm{p}_{A}(\bm{q})-\bm{p}_{B}(\bm{q})\|_2.
\label{eq:workspace_boundary_barriers}
\end{equation}
Here, $d_{\max}>0$ is the maximum allowable distance.
Let
\begin{equation}
\hat{\bm{n}}_{\mathrm{WS}}(\bm{q})
\triangleq
\frac{\bm{p}_{A}(\bm{q})-\bm{p}_{B}(\bm{q})}{\|\bm{p}_{A}(\bm{q})-\bm{p}_{B}(\bm{q})\|_2},
\label{eq:workspace_normal}
\end{equation}
and
\begin{equation}
\bm{J}_{\mathrm{WS}}(\bm{q})
=
-\hat{\bm{n}}_{\mathrm{WS}}(\bm{q})^\top\!\left(\bm{J}_{A}(\bm{q})-\bm{J}_{B}(\bm{q})\right).
\label{eq:workspace_signed_jacobian}
\end{equation}
Then,~\eqref{eq:issf_cbf_constraint_template} yields
\begin{equation}
\bm{J}_{\mathrm{WS}}(\bm{q})\dot{\bm{q}}
\ge
-\alpha^{\mathrm{WS}}\!h^{\mathrm{WS}}_{\text{AB}}(\bm{q})
+ \frac{1}{\epsilon^{\mathrm{WS}}}
\left\|\bm{J}_{\mathrm{WS}}(\bm{q})\right\|^2.
\label{eq:workspace_qdot_maxnorm}
\end{equation}
\begin{remark}
When multiple CBFs are imposed, the resulting constraints may conflict with one another and lead to an infeasible QP. On the humanoid robot platform used in this study, infeasible QPs were not observed in practice because the upper body provides substantial redundancy for task execution. For robot platforms with fewer available DoFs, the QP can be relaxed by introducing slack variables. This often yields a practical solution, although it may no longer guarantee safety in general.
\end{remark}

\section{RESULTS OF SIMULATIONS AND EXPERIMENTS} \label{sec:result}

This section presents simulation and real-robot experiments to validate safety transfer from the kinematic level to full-order humanoid dynamics using the proposed ISSf-CBF filter. 
The simulations evaluate whether the proposed ISSf-CBF filter can ensure collision-free behavior during hand-trajectory tracking and locomotion scenarios in the presence of model mismatch, and compare its collision-avoidance performance with baseline safety filters across different parameter settings.
The real-robot experiments evaluate whether the proposed framework maintains stable single-leg balance under contact wrench cone constraints and enforces workspace-boundary constraints for both hands during TAICHI motions.
The real-robot experiments further assess whether the proposed safety filter enforces obstacle avoidance during teleoperation while preserving the user's intended task execution.

\subsection{System overview}
This subsection provides a system overview for both simulation and real-robot experiments.
The humanoid robot TOCABI~\cite{schwartz2022tocabi} used in this study has 33 degrees of freedom (DoF): 16 in the arms, 12 in the legs, 3 in the waist, and 2 in the neck.
Its physical dimensions are approximately 1.8~m in height and 100~kg in mass, with each foot measuring 15~cm in width and 30~cm in length.
The upper-body actuators use Parker BLDC motors with harmonic drives, whereas the lower-body actuators use Kollmorgen BLDC motors with harmonic drives.
Current control is implemented using Elmo Motion Control's Gold Solo Whistle servo controllers, and communication between the servo controllers and the main PC is established via EtherCAT.
The main PC is equipped with a 3.7 GHz octa-core processor and 16 GB of RAM.
A MicroStrain 3DM-GX5-25 IMU mounted on the pelvis and ATI Mini85 F/T sensors mounted on each ankle are used for floating-base pose estimation~\cite{henze2016passivity}.
Both the control algorithm and the torque command loop run at 2 kHz.
The simulator employed in this study is MuJoCo~\cite{todorov2012mujoco}.
The QP problems in \eqref{eq:issf_qp_ki} and \eqref{eq:dynwbc_qp_obj} are solved using qpOASES~\cite{ferreau2014qpoases}.
At each 2~kHz control cycle, the ISSf-CBF QP in \eqref{eq:issf_qp_ki} is solved in the main thread, whereas the DynWBC QP in \eqref{eq:dynwbc_qp_obj} is solved in a parallel thread.
RBDL~\cite{felis2017rbdl} is used for robot kinematics and dynamics computations.

\subsection{Simulations to validate safety transfer from ROM to FOM}
Simulations were conducted to evaluate whether the proposed method can transfer kinematic-level safety to the full-order humanoid dynamics model more effectively than baseline methods and to examine how the ISSf property improves safety enforcement in the presence of model mismatch.
The baseline methods used for comparison are summarized as follows:
\begin{itemize}
    \item \textit{w/o-CBF:} KinWBC and DynWBC are used without any safety filter.
    \item \textit{CBF:} standard CBF constraints are enforced without the ISSf term, i.e., $\frac{1}{\epsilon}\|\nabla h(\bm{q})\|^2$.
    \item \textit{eCBF:} eCBF constraints are enforced only at the acceleration level in DynWBC.
    \item \textit{ISSf-CBF (proposed):} the safety filter in Section~\ref{sec:safety_filter} is used.
\end{itemize}
To incorporate the eCBF constraints in DynWBC, the exponential barrier function is defined as follows~\cite{nguyen2022exponential, khazoom2022collision}:
\begin{equation}
h_e(\bm{q},\dot{\bm{q}})=\dot{h}(\bm{q},\dot{\bm{q}})+\alpha h(\bm{q}),
\label{eq:ecbf_he_definition}
\end{equation}
and the following condition is enforced:
\begin{equation}
\dot{h}_e(\bm{q},\dot{\bm{q}},\ddot{\bm{q}})
=
\frac{\partial h_e(\bm{q},\dot{\bm{q}})}{\partial \bm{q}}\dot{\bm{q}}
+
\frac{\partial h_e(\bm{q},\dot{\bm{q}})}{\partial \dot{\bm{q}}}\ddot{\bm{q}}
\ge
-\alpha_e h_e(\bm{q},\dot{\bm{q}}).
\label{eq:ecbf_he_constraint}
\end{equation}
Since the eCBF condition is affine in $\ddot{\bm{q}}$, eCBF-based safety constraints are incorporated into DynWBC as linear inequality constraints.
Following the experimental setup in~\cite{khazoom2022collision}, the parameter $\alpha_e$ was set to $\alpha$.


To incorporate model mismatch into the simulation, the robot URDF model extracted from the CAD files was used for kinematics and dynamics computations in RBDL, whereas the MuJoCo XML model was used to define the simulated robot dynamics and its interaction with the environment. Specifically, the mismatch was introduced only in the MuJoCo XML model by increasing the link masses by $20\%$. Because this mass scaling was not reflected in the URDF model, the controller performed kinematics and dynamics computations using the nominal model, thereby emulating model uncertainty in real-robot systems.



\begin{figure*}[!t]
    \centering
    \includegraphics[width=\textwidth]{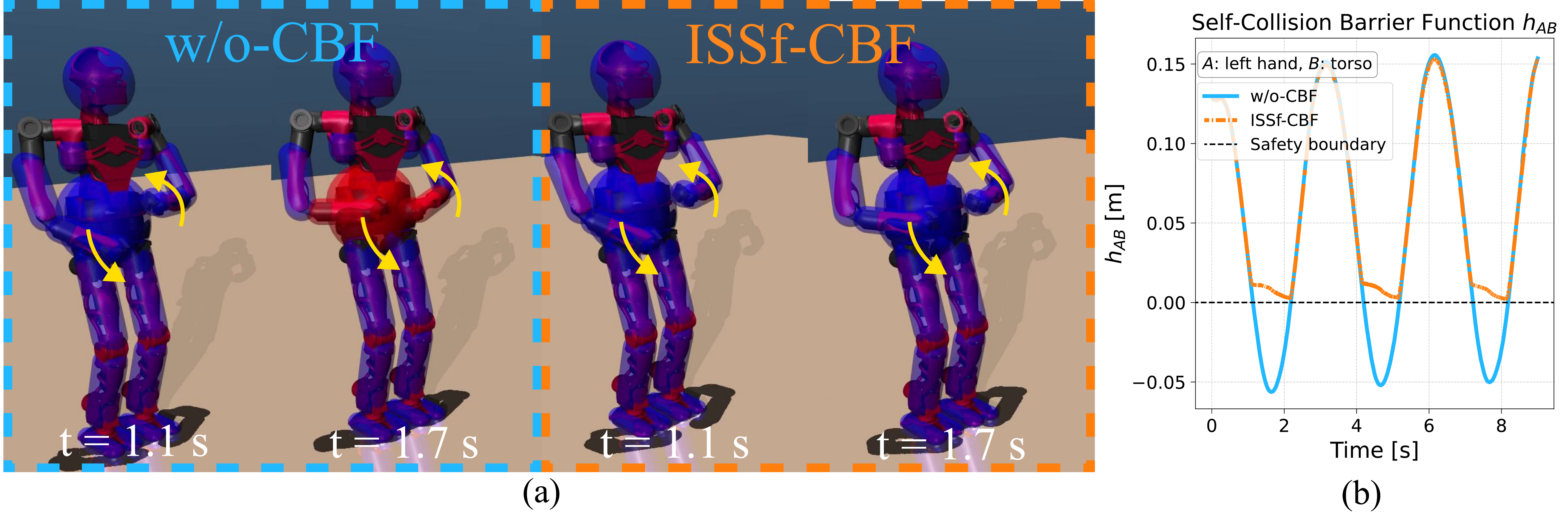}
    \caption{Snapshots of the hand-trajectory-tracking scenario and signed-distance barrier results. 
    The geometries are defined for collision avoidance; blue indicates that no collision has occurred, whereas red indicates that a collision has occurred. In (a), 
    the \textit{w/o-CBF} case shows collisions, whereas the \textit{ISSf-CBF} case remains collision-free. In (b), the signed-distance barrier function between the left hand and the torso is shown.}
    \label{fig:min_distance}
\end{figure*}

\begin{figure*}[!t]
    \centering
    \includegraphics[width=\textwidth]{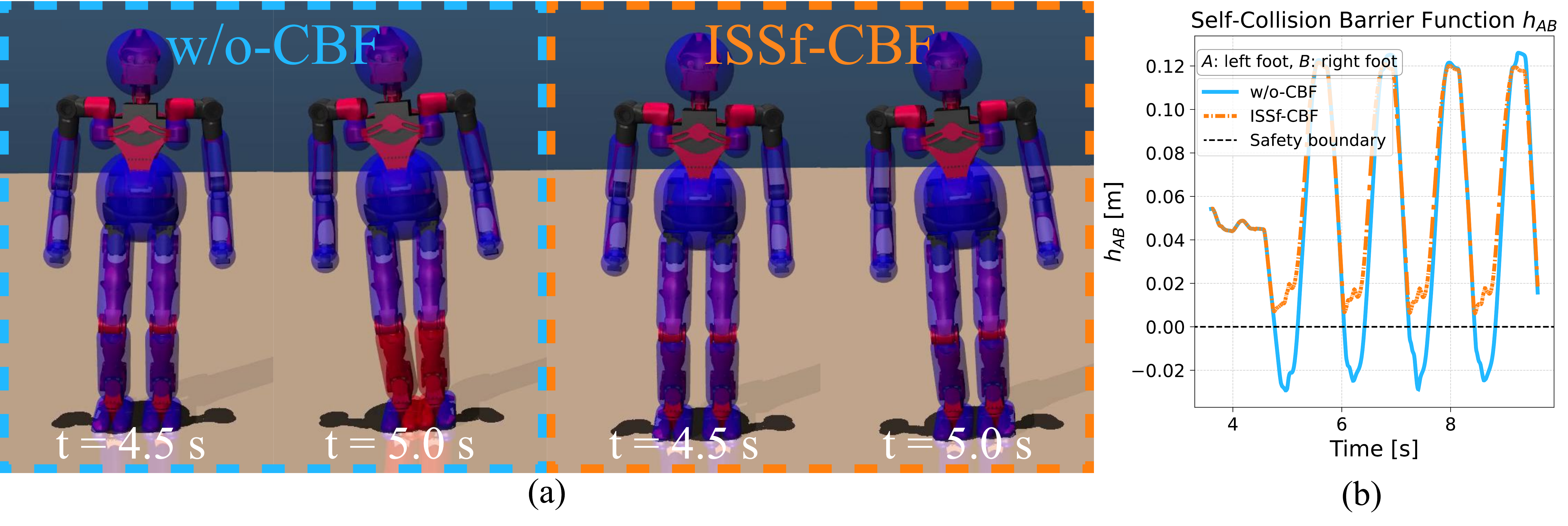}
    \caption{Snapshots of the locomotion scenario and signed-distance barrier results. 
    The geometries are defined for collision avoidance; blue indicates that no collision has occurred, whereas red indicates that a collision has occurred. In (a),
    the \textit{w/o-CBF} case shows collisions, whereas the \textit{ISSf-CBF} case remains collision-free. In (b), the signed-distance barrier function between the left and right feet is shown.}
    \label{fig:foot_snapshot}
\end{figure*}

\subsubsection{Simulations to validate safety assurance by the proposed \textit{ISSf-CBF} filter under model mismatch}
Comparative simulations between \textit{ISSf-CBF} and \textit{w/o-CBF} were conducted to verify whether the proposed filter can consistently prevent self-collision between robot links in hand-trajectory-tracking and locomotion scenarios in the presence of model mismatch.
In the hand-trajectory-tracking scenario, both hands tracked 3D circular trajectories that would nominally collide with the torso.
Fig.~\ref{fig:min_distance}(a) shows snapshots of the hand-trajectory-tracking scenario, and Fig.~\ref{fig:min_distance}(b) shows the signed-distance barrier function between the left hand and the torso.
In the \textit{w/o-CBF} case, the left hand penetrated the torso by more than $5$~cm, as indicated by the negative signed-distance barrier value.
In contrast, the proposed \textit{ISSf-CBF} filter kept the signed-distance barrier value positive throughout the hand-trajectory-tracking scenario, indicating that self-collisions between the hand and the torso were avoided.

In the locomotion scenario, the robot performed lateral steps that would nominally lead to foot-to-foot collisions.
Fig.~\ref{fig:foot_snapshot}(a) shows snapshots of the locomotion scenario, and Fig.~\ref{fig:foot_snapshot}(b) shows the signed-distance barrier function between the left and right feet.
In the \textit{w/o-CBF} case, the feet penetrated each other by more than $3$~cm, as indicated by the negative signed-distance barrier value.
In contrast, the proposed \textit{ISSf-CBF} filter kept the signed-distance barrier value positive throughout the locomotion scenario, indicating that self-collisions between the feet were avoided.
These results demonstrate that the proposed \textit{ISSf-CBF} filter effectively transfers the kinematic safety constraints in~\eqref{eq:issf_qp_ki} to the FOM controller in~\eqref{eq:motor_torque_command}, even in the presence of model mismatch.

\subsubsection{Simulations to compare safety assurance performance with baseline methods under CBF parameter variations} \label{sec:result-baseline}
The performance of all baseline methods was compared in the aforementioned hand-trajectory-tracking scenario by varying the CBF parameter $\alpha_{AB}^{\mathrm{SC}}\in\{1,5,10,\cdots,50\}\,\mathrm{s}^{-1}$ and the ISSf parameter $\epsilon_{AB}^{\mathrm{SC}}\in\{10,20,30\}$.
\begin{figure}[!t]
    \centering
    \includegraphics[width=\columnwidth]{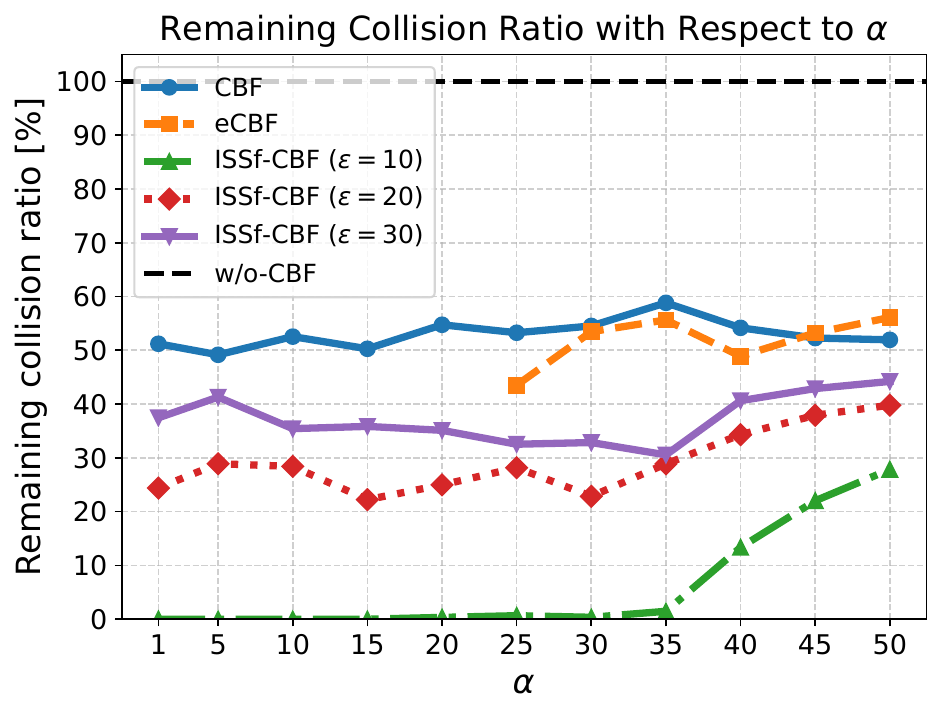}
    \caption{Performance comparison of baseline methods in the hand-trajectory-tracking scenario under different $\alpha_{AB}^{\mathrm{SC}}$ and $\epsilon_{AB}^{\mathrm{SC}}$ settings.}
    \label{fig:col_ratio}
\end{figure}
Fig.~\ref{fig:col_ratio} shows the comparative performance trends across parameter settings.
Here, the remaining collision ratio is defined as the ratio of the number of collisions that still occur, i.e., those for which $h_{AB}^{\mathrm{SC}}(\bm{q}) < 0$, after applying a safety filter to the total number of collisions in the reference hand trajectory.

For the \textit{CBF} baseline, the collision ratio remained around 50\% across the tested values of $\alpha$; that is, varying $\alpha$ did not lead to an improvement in the collision ratio in the present scenario.
Moreover, when $\alpha\geq 30$, abrupt velocity changes near the constraint boundary induced severe jitter in the joint-velocity command. 
For the \textit{eCBF} baseline, cases with $\alpha\leq 20$ were excluded because the resulting motion deviated substantially from the reference hand trajectory in that regime.
This behavior is consistent with the increased tracking error at low $\alpha$ reported in~\cite{khazoom2022collision}.
Even in the evaluated range, the \textit{eCBF} baseline showed an average collision ratio of about 50\%; that is, varying $\alpha$ did not lead to an improvement in the collision ratio in the present scenario.
In contrast, the proposed \textit{ISSf-CBF} method remained collision-free when both $\epsilon$ and $\alpha$ were set to relatively low values.
For \textit{ISSf-CBF} with $\epsilon=10$, no collision was observed for $\alpha\leq 30$; however, when $\alpha\geq 35$, the remaining collision ratio increased due to severe jitter in the safety-velocity commands near constraint boundaries.
As $\epsilon$ increased, the collision ratio also increased, indicating that the selected parameters were insufficient to compensate for bounded ROM--FOM disturbances.

Overall, compared with the evaluated baselines, the \textit{ISSf-CBF}-based safety filter provides superior safety assurance under bounded ROM--FOM disturbances even in the presence of model mismatch.
Moreover, these results experimentally support the claim in~\cite{cohen2025safety} that setting $\alpha$ and $\epsilon$ to relatively low values can enable effective safety transfer from ROM to FOM.

\subsection{Real-robot experiments to validate safety transfer from ROM to FOM}
Real-robot experiments were conducted on the humanoid robot TOCABI to evaluate whether the proposed \textit{ISSf-CBF}-based safety filter can be deployed in real time while robustly enforcing safety constraints under discrepancies between the nominal model and the actual humanoid hardware.
For example, the nominal URDF model used for control estimated the total robot mass as 96 kg, whereas the actual robot mass measured on a scale was 102 kg, indicating an approximately 6 \% model discrepancy.
The experiments consist of single-leg balancing with hand control and teleoperation, validating contact stability, workspace-boundary enforcement, and obstacle avoidance on the real robot.
For teleoperation, the retargeting algorithm in~\cite{bertrand2024teleop} was used with a PICO 4 Ultra VR device. The trackers measured the human ankle and wrist poses, as well as the chest and head orientations; these measurements were retargeted to the robot, and KinWBC in~\eqref{eq:kinwbc_qdot} generated the corresponding robot commands.
Videos of the experiments can be found on our project webpage.

\subsubsection{Real-robot experiment to validate contact stability and workspace-boundary enforcement during single-leg balancing with hand control}
\begin{figure*}[!t]
    \centering
    \includegraphics[width=\textwidth]{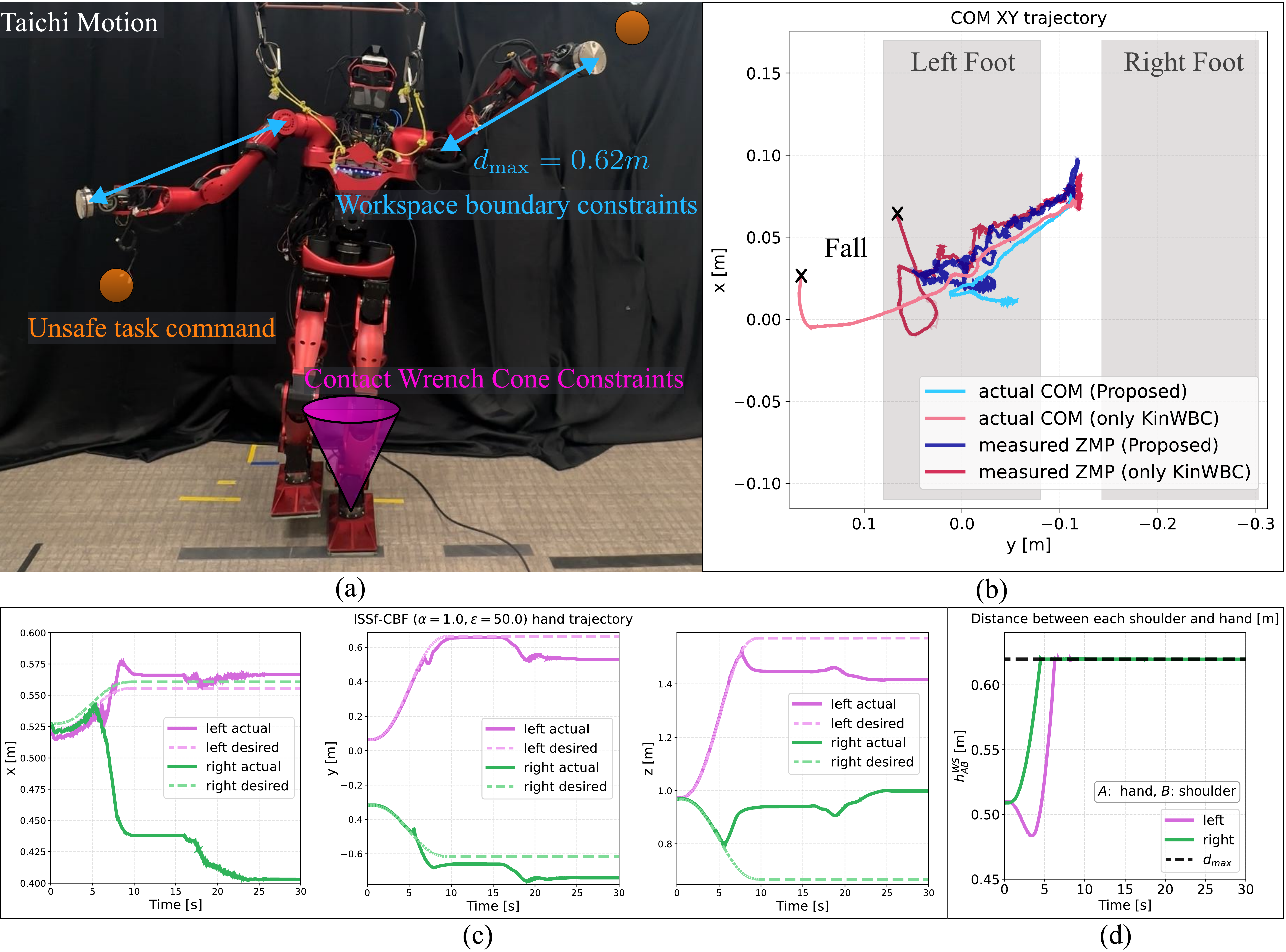}
    \caption{Real-robot validation of contact stability and workspace-boundary enforcement during single-leg balancing with hand control. The experimental snapshot is shown in (a), and the output data are shown in (b)--(d).}
    \label{fig:taichi}
\end{figure*}
This experiment verifies whether the proposed hierarchical control framework can simultaneously maintain contact stability and enforce workspace-boundary constraints for both hands during single-leg balancing.
The single-leg balancing motion with hand control, referred to as TAICHI, was constructed as follows:
During the initial 10 $\mathrm{s}$, the robot's center of mass~(CoM) moved toward the center of the left foot. 
At the same time, the right hand was commanded to move 30 cm in the negative $y$- and $z$-directions from its initial position, while the left hand was commanded to move 60 cm in the positive $y$- and $z$-directions from its initial position.
The reference trajectories for both hands were generated using quintic splines. 
Without the safety filter, KinWBC may drive the robot state toward a kinematic singularity when solving inverse kinematics for unreachable workspace targets, resulting in excessive joint-velocity commands and potentially unsafe motions.
Therefore, the maximum allowable distance $d_{\max}$ between each shoulder and its corresponding hand was selected to prevent the robot from approaching singular configurations. For the humanoid robot used in this experiment, $d_{\max}$ was set to 62 cm.
To maintain stable balance, the Zero Moment Point (ZMP)~\cite{vukobratovic2004zero} trajectory was planned to coincide with the CoM trajectory. 
In addition, a desired contact force $\bm{F}_c^{\text{des}}$ for balance control was generated using CoM trajectory feedback and a contact wrench distributor~\cite{kajita2010biped}.
DynWBC then computed whole-body torque commands $\bm{\tau}^{\text{opt}}$ to track the desired contact wrenches while explicitly enforcing contact wrench cone constraints~\eqref{eq:dynwbc_qp_fric}.
Finally, the robot was commanded to lift the right foot by 15 cm in the positive $z$-direction for 10 s.

Fig.~\ref{fig:taichi}(a) shows that the robot maintained balance during single-leg support while both arms satisfied the workspace-boundary constraints. Fig.~\ref{fig:taichi}(b) shows the desired CoM trajectory, the measured CoM, and the global ZMP computed from the net contact wrench measured by the F/T sensors attached to both feet, all projected onto the $xy$ plane. 
The measured ZMP remained inside the support polygon of the left foot during single-leg balance control. In contrast, the robot fell when the torque command in~\eqref{eq:motor_torque_command} was applied without $\bm{\tau}^{\mathrm{opt}}$, i.e., when only the joint PD feedback terms were used.
This comparison shows that desired contact wrench tracking and contact wrench cone constraints in DynWBC are critical for maintaining balance stability in humanoid robots.
Fig.~\ref{fig:taichi}(c) and (d) show the three-dimensional trajectories of both hands and the relative distances between each shoulder and its corresponding hand, respectively. 
Before reaching the workspace boundary, both hands accurately tracked the reference trajectories. 
In contrast, when the relative distance between each hand and its corresponding shoulder approached the maximum allowable distance $d_{\max}=62\,\mathrm{cm}$, the safety filter relaxed trajectory tracking and modified the hand commands so that both hands remained within the admissible workspace.
As a result, the relative distances between the hands and their corresponding shoulders remained below $d_{\max}$ throughout the experiment, confirming that the workspace-boundary constraints were satisfied.

These results demonstrate that the proposed ISSf-CBF-based hierarchical framework can leverage whole-body inverse-dynamics control to maintain contact stability while preserving task performance and kinematic safety under real-hardware uncertainties.

\subsubsection{Real-robot experiment to validate obstacle-avoidance enforcement during teleoperation}

\begin{figure*}[!t]
    \centering
    \includegraphics[width=\textwidth]{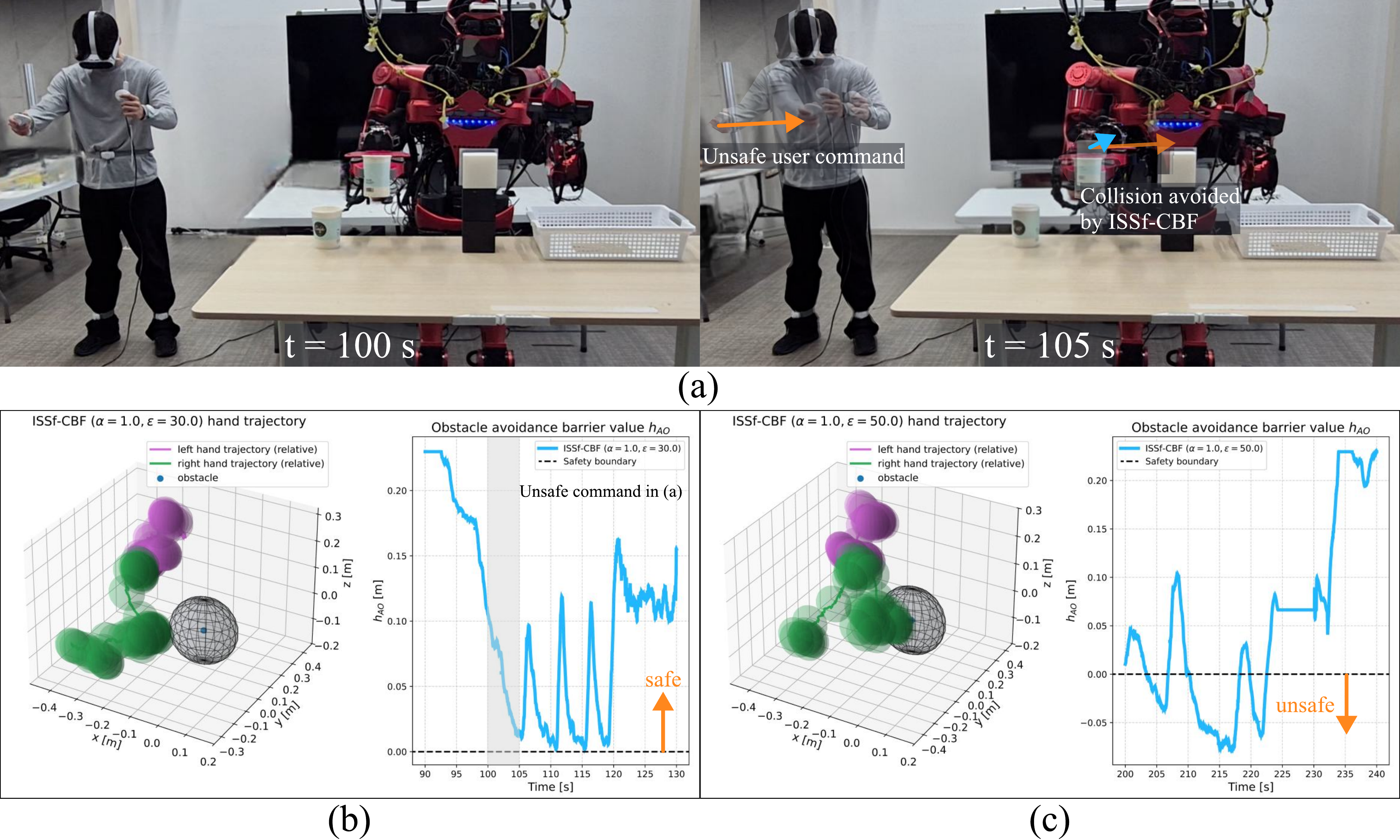}
    \caption{Real-robot validation of obstacle-avoidance enforcement during teleoperation. 
    The experimental snapshots are shown in (a), and the output data for two \textit{ISSf-CBF} parameter settings, $\varepsilon_{AO}^{OC}=30$ and $\varepsilon_{AO}^{OC}=50$, are shown in (b) and (c), respectively.}
    \label{fig:real_snapshot}
\end{figure*}

The objective of this experiment is to evaluate the obstacle-avoidance performance of the proposed \textit{ISSf-CBF} in a teleoperated manipulation task.
Under teleoperation, the user was asked to pick up two paper cups initially placed on the left side of the table and place them into a basket located on the right side.
An external obstacle was placed between the paper cups and the basket, creating a risk of collision between the robot's hand link and the obstacle, depending on the user's commands.
Therefore, this experiment examines whether the proposed \textit{ISSf-CBF} can effectively avoid the obstacle while preserving the user's intended motion as much as possible.
During the experiment, the center position of the external obstacle, $\bm{p}_{O}^{r}$, was estimated using QR markers, and its velocity, $\bm{v}_{O}$, was estimated using a linear Kalman filter~\cite{przybyla2017detection}. 
The obstacle used in the experiment was a rectangular box with a width of 10~cm, a depth of 10~cm, and a height of 6~cm. The external obstacle was modeled as a sphere with a radius of 11.5~cm, and the collision geometry of the robot hand was modeled as a sphere with a radius of 6~cm.

Fig.~\ref{fig:real_snapshot}(a) presents experimental snapshots.
The snapshots illustrate that, when the user commanded the right hand toward the obstacle (orange arrow), the \textit{ISSf-CBF} safety filter modified the nominal motion reference generated from the user command so that the right-hand collision geometry remained within the collision-free set (blue arrow), thereby preventing a collision between the obstacle and the right-hand link.
Fig.~\ref{fig:real_snapshot}(b) and (c) show the output data obtained with $\varepsilon_{AO}^{OC}=30.0$ and $\varepsilon_{AO}^{OC}=50.0$, respectively, including the collision-geometry trajectories of the two hands and the obstacle-avoidance barrier value $h_{AO}$.

In Fig.~\ref{fig:real_snapshot}(b), the user provided a command toward the obstacle at around 100~s, and the obstacle-avoidance barrier value approached zero.
At that time, the \textit{ISSf-CBF} safety filter kept the right-hand link in the region satisfying $h_{AO} \ge 0$.
The same motion was then repeated three additional times, and in all cases the \textit{ISSf-CBF} safety filter modified the nominal motion reference generated from the user command such that no collision with the obstacle occurred.
In Fig.~\ref{fig:real_snapshot}(c), the user provided a command toward the obstacle at around 200~s.
Across the three trials, the obstacle-avoidance barrier value fell below zero in every case, and a maximum penetration depth of 8~cm was observed between the collision geometries of the obstacle and the right-hand link.

These results demonstrate that, with an appropriate choice of $\varepsilon$, the \textit{ISSf-CBF}-based safety filter can effectively ensure safety under real-robot uncertainties.

\section{CONCLUSION} \label{sec:conclusion}
This paper presents a hierarchical safety-critical whole-body control framework for humanoid robots based on ISSf-CBFs.
The proposed framework integrates KinWBC, the ISSf-CBF-based safety filter, and DynWBC to robustly enforce kinematic safety constraints under unknown disturbances.
To transfer kinematic-level safety guarantees to full-order humanoid dynamics, safety constraints are formulated on a whole-body velocity kinematic model, and the ISSf-CBF parameters are conservatively tuned to account for unknown disturbances.
The proposed method was validated through simulations and real-robot experiments involving locomotion, teleoperation, and single-leg balancing with hand control.
The results show improved safety margins under model mismatch and reliable enforcement of contact stability and kinematic safety constraints in real time.

In the proposed framework, contact stability is enforced through contact wrench constraints, and the robot may still fail to maintain balance under uncertainties in terrain height or friction conditions. 
Recent reinforcement-learning-based locomotion and whole-body control methods have demonstrated strong robustness by learning balance-recovery behaviors through extensive trial and error in simulation~\cite{zhang2025hub}. 
Moreover, emerging studies that combine CBFs with reinforcement learning indicate a promising direction for incorporating safety-critical filtering into policy learning and execution~\cite{yang2025cbf}. 

Building on this trend, our future work will investigate how the proposed ISSf-CBF-based safety filter can be integrated with reinforcement-learning-based locomotion and whole-body control policies to improve robustness while maintaining explicit safety guarantees.

\end{document}